\documentclass[10pt]{article}

\usepackage{lineno}
\modulolinenumbers[5]

\usepackage{wrapfig}

\usepackage{amsmath, amssymb}
\usepackage[all]{xy}
\usepackage{graphicx}

\begin{document}


\title{Uncertain Linear Logic via Fibring of Probabilistic and Fuzzy Logic}

\author{Ben Goertzel}





\maketitle

\begin{abstract}
Beginning with a simple semantics for propositions, based on counting observations, it is shown that probabilistic and fuzzy logic correspond to two different heuristic assumptions regarding the combination of propositions whose evidence bases are not currently available.   These two different heuristic assumptions lead to two different sets of formulas for propagating quantitative truth values through lattice operations.  It is shown that these two sets of formulas provide a natural grounding for the multiplicative and additive operator-sets in linear logic.   The standard rules of linear logic then emerge as consequences of the underlying semantics.   The concept of linear logic as a ``logic of resources" is manifested here via the principle of ``conservation of evidence" -- the restrictions to weakening and contraction in linear logic serve to avoid double-counting of evidence (beyond any double-counting incurred via use of heuristic truth value functions).
\end{abstract}




\section{Introduction}

Linear logic  \cite{girard1987linear} comprises a rich and fascinating formal system that summarizes, in a nuanced way, the way logical inference works if one treats the pool of potential premises of inferences as a resource to be meted out and accounted for.    The linear logic abstractions can be applied to practical reasoning systems in a variety of different ways, and can be grounded in concrete domain-specific inference formalisms via multiple routes as well.

Here we connect linear logic to uncertain reasoning based on observational semantics.   Beginning with a simple semantics for propositions, based on counting observations, we argue that probabilistic and fuzzy logic correspond to two different heuristic assumptions regarding the combination of propositions whose evidence bases are not currently available.   These two different heuristic assumptions lead to two different sets of formulas for propagating quantitative truth values through lattice operations.  Given this set-up, it becomes immediately apparents that these two sets of formulas instantiate the same algebraic and conceptual relationships as the multiplicative and additive operator-sets in linear logic.   The standard rules of linear logic then emerge as consequences of the underlying semantics of fuzzy and probabilistic evidence management.   

The concept of linear logic as a ``logic of resources" is manifested here via the principle of ``conservation of evidence" -- the restrictions to weakening and contraction in linear logic serve to avoid double-counting of evidence (beyond any double-counting incurred via use of heuristic truth value functions).

\section{Core Concepts of Linear Logic}

First we summarize some basic concepts of linear logic \cite{girard1987linear}.

In linear logic, every propositional variable is considered as a proposition.

For each proposition A, there is a proposition $A^\perp$, the negation of A.

For each proposition A and proposition B, there are four additional propositions:

\begin{itemize}
\item $ A \& B$ (read ``with"), the additive conjunction of And B;
\item  $A \oplus B$ (read ``plus"), the additive disjunction of And B;
\item  $A \otimes B$ (read ``times"), the multiplicative conjunction of And B;
\item  $A \mid B$, the multiplicative disjunction of And B.
 \end{itemize}

There are also four constants to go with these four binary operations:

\begin{itemize}
\item  $\top $(read ``top"), the additive truth;
 \item $\mathbf{0} $(read ``zero"), the additive falsity;
\item $ \mathbf{1} $(read ``one"), the multiplicative truth;
\item $ \bot $(read ``bottom"), the multiplicative falsity.
 \end{itemize}
 
; and, for each proposition A, there are two additional propositions:

\begin{itemize}
 \item !{A} (read ``of course"), the exponential conjunction of A;
\item  ?{A} (read ``why not"), the exponential disjunction of A.
 \end{itemize}
 
 The interpretation of these operations is a complex issue on which there are many
 complementary perspectives. \cite{lincoln1992operational} gives a concrete interpretation in terms of a specific
 computational model .   An informal ``resource interpretation" is commonly
 discussed, for instance with a ``vending machine" metaphor \cite{hoare1983communicating}:
 
 {\it
 Suppose we represent having a candy bar by the atomic proposition $\textrm{candy}$, and having a dollar by \$1. To state the fact that a dollar will buy you one candy bar, we might write the implication $ \$1 \rightarrow \textrm{candy}$. But in ordinary (classical or intuitionistic) logic, from And $A \rightarrow B$ one can conclude $A \wedge B$. So, ordinary logic leads us to believe that we can buy the candy bar and keep our dollar! Of course, we can avoid this problem by using more sophisticated encodings...}
 
 The vending machine is used as a metaphor for linear logic as follows:
 
 {\it ... rather than $ \$1 \rightarrow \textrm{candy} $, we express the property of the vending machine as a linear implication $ \$1 \rightarrow candy$. From \$1 and this fact, we can conclude candy, but not $ \$1 \otimes candy$.  In general, we can use the linear logic proposition $A \rightarrow B$ to express the validity of transforming resource A into resource B.
 
Multiplicative conjunction ($A \otimes B$) denotes simultaneous occurrence of resources, to be used as the consumer directs. For example, if you buy a stick of gum and a bottle of soft drink, then you are requesting $\textrm{gum} \otimes \textrm{drink}$. 

Additive conjunction ($A \& B$) represents alternative occurrence of resources, the choice of which the consumer controls. If in the vending machine there is a packet of chips, a candy bar, and a can of soft drink, each costing one dollar, then for that price you can buy exactly one of these products. Thus we write $ \$1 \rightarrow(\textrm{candy} \& \textrm{chips} \& \textrm{drink})$. We do not write $ \$1 \rightarrow (\textrm{candy} \otimes \textrm{chips} \otimes \textrm{drink})$, which would imply that one dollar suffices for buying all three products together. However, from $ \$1 \rightarrow (\textrm{candy} \& \textrm{chips} \& \textrm{drink})$, we can correctly deduce $ \$3 \rightarrow (\textrm{candy} \otimes \textrm{chips} \otimes \textrm{drink})$, where $ \$3 = \$1 \otimes \$1 \otimes \$1$. 

Additive disjunction ($A \oplus B$) represents alternative occurrence of resources, the choice of which the machine controls. For example, suppose the vending machine permits gambling: insert a dollar and the machine may dispense a candy bar, a packet of chips, or a soft drink. We can express this situation as $ \$1 \rightarrow (\textrm{candy} \oplus \textrm{chips} \oplus \textrm{drink} )$. 

Multiplicative disjunction ($A \mid B$) is more difficult to gloss in terms of the resource interpretation.

}

\noindent Overall, the metaphor somewhat works, but then gets confusing when pushed too far.  The approach taken here grounds linear logic operators in {\it evidential resources}, which is exact rather than metaphorical.

\section{Observational Semantics for Uncertain Inference}

Following the approach outlined in our prior writings on Probabilistic Logic Networks \cite{PLN}, we propose to ground the semantics of propositions in finite sets of observations  made by a particular system.
 
Suppose each proposition $A$ under consideration is supported by a certain set $O_A$ of observations, and has a certain quantitative truth value (which may be a single number of a tuple of numbers; we will consider, specifically, pair truth values below).
 
 To calculate the probabilistic truth value of $A \wedge B$ (the conjunction of And B) or $A \vee B$ (the disjunction of And B), if one has not retained
 the observation-sets $O_A$ and $O_B$, one has to make some assumptions about the relationship between $O_A$ and $O_B$.
 
 Two assumptions are particularly simple to make:
 
 \begin{itemize}
 \item {\bf Max Overlap}: That $O_A$ and $O_B$ maximally overlap: i.e. if they are the same size, then they are identical ... and if they are different sizes, then one is entirely a subset of the other
 \item {\bf Independence}: That $O_A$ and $O_B$ are probabilistically independent samples from some larger space
 \end{itemize}
 
 These two assumptions lead to different uncertain truth value formulas
 
  \begin{itemize}
 \item {\bf Max Overlap}: 
 \begin{itemize}
 \item $p( A \wedge B ) = min( p(A), p(B) )$
 \item  $p( A \vee B ) = max( p(A) , p(B) ) $
 \end{itemize}
 \item {\bf Independence}: 
   \begin{itemize}
 \item $p( A \wedge B ) = p(A) * p(B)$
 \item  $p( A \vee B ) = p(A) + p(B) - p(A) * p(B) $
 \end{itemize}
 \end{itemize}
 
\noindent It happens that these are both t-norms, meaning they have intuitively natural algebraic symmetry properties \cite{gupta1991theory}.

Alongside $p(A)$, it is worthwhile to keep track of $n(A)$, the number of observations on which $p(A)$ is based.   This yields two-component truth values 
$(s,n)$, which are a variant of what are called ``simple truth values'' in PLN.  In this regard the two assumptions yield different formulae as well:

 \begin{itemize}
 \item {\bf Max Overlap}: 
 \begin{itemize}
 \item $n( A \wedge B ) = min( n(A), n(B) )$
 \item  $n( A \vee B ) = max( n(A) , n(B) ) $
 \end{itemize}
 \item {\bf Independence}: 
   \begin{itemize}
 \item $n( A \wedge B ) = n(A) * n(B)$
 \item  $n( A \vee B ) = n(A) + n(B) - n(A) * n(B) $
 \end{itemize}
 \end{itemize}
 
 As a concrete example, suppose one is evaluating a population of $100$ people, and $A$ = American while $B$ =  crazy.   Suppose we know
 
 \begin{itemize}
 \item n(A) = 20, i.e. 20 people were observed to evaluate the odds of a person being American
 \item p(A) = .5, i.e. based on these 20 evaluations, half were observed to be American and the other half not
 \item n(B) = 10, i.e. 10 people were observed to evaluate the odds of a person being crazy
 \item p(B) = .3, i.e. 3 of these 10 people were observed to be crazy, the others not
  \end{itemize}
 
\noindent Then we are considering two different situations:
 
  \begin{itemize}
 \item {\bf Max Overlap}: The 10 people evaluated regarding craziness, were a subset of the 20 people evaluated regarding American-ness
 \item {\bf Independence}: The 20 people evaluated regarding craziness, and the 10 people evaluated regarding American-ness, were independently randomly selected as subsets of the original 100 people
 \end{itemize}
 
\noindent In these two cases we have:
 
   \begin{itemize}
 \item {\bf Max Overlap}: 
  \begin{itemize}
 \item $p( A \wedge B ) = .3$, based on looking at the 10 people in $O_A \wedge O_B$
 \item  $p( A \vee B ) = .5 $, since in this assumption all the crazy people evaluated were also American ... in this assumption the evidence for non-American crazy people is zero
  \item $n( A \wedge B ) = 10$, the group of people evaluated for both American-ness and craziness
 \item  $n( A \vee B ) = 20$, the group of people evaluated for American-ness (all of whom were also evaluated for craziness)
 \end{itemize}
 \item {\bf Independence}: 
  \begin{itemize}
 \item $p( A \wedge B ) = .5 * .3 = .15$
 \item  $p( A \vee B ) = .5 + .3 - .15 = .65 $
  \item $n( A \wedge B ) = 20 * 10 / 100 = 2$, the expected amount of overlap between the 20 randomly selected people evaluated for American-ness and the 10 randomly selected people evaluated for craziness
 \item  $n( A \vee B ) = 28 $, the expected number of people in the union of the 20 randomly selected people evaluated for American-ness and the 10 randomly selected people evaluated for craziness
 \end{itemize}
 \end{itemize}
 
 \section{From Observational Semantics to Linear Logic}
 
 The core novel suggestion I want to make here is to interpret:
 
 \begin{itemize}
\item $ A \& B$ = conjunction of $A$ and $B$ according to the assumption of Max Overlap of the underlying evidence sets
\item  $A \oplus B$ = disjunction of $A$ and $B$ according to the assumption of Max Overlap of the underlying evidence sets
\item  $A \otimes B$ = conjunction of $A$ and $B$ according to the assumption of probabilistic independence of the underlying evidence sets
\item  $A \mid B$ = disjunction of $A$ and $B$ according to the assumption of probabilistic independence of the underlying evidence sets
 \end{itemize}

This can be viewed as a different sort of ``resource interpretation" of linear logic, in which the resources involved are not candy bars and such but rather observations made (evidence gathered) in favor of propositions.   More explicitly, in this interpretation/construction of linear logic:

\begin{itemize}
\item Multiplicative conjunction represents utilization of a body of potential observations (e.g. the 100 people potentially observable in the example above) to evaluate two properties independently and concurrently.   The set of pairs (possible observation giving evidence about American-ness, possible observation giving evidence about craziness) is quite large, since each of these components of the pair is selected independently from the whole body of observations (100 in the example).
\item Additive conjunction represents minimalist utilization of a body of potential observations, for making a combined observation of two properties.   We assume the making of as few observations as possible, consistent with the known data.
\item Additive disjunction represents minimalist use of evidence items (for making an observation of one or the other of two properties): a certain piece of evidence might be used for evaluating craziness only, or for evaluating both craziness and American-ness
\item Multiplicative disjunction represents use of evidence items for evaluating one or the other of two properties, in a way that assumes the processes of evaluating these properties are independent.  A certain piece of evidence might be used for evaluating craziness only, for evaluating American-ness only, or for evaluating both.
\end{itemize}

The $0$ and $1$ constants of linear logic may then be interpreted as follows:

\begin{itemize}
\item  $\top $(read ``top"), the additive truth, represents a set of observations whose size is the minimum needed to be logically compatible with all the known information about counts and probabilities
\item $ \mathbf{1} $(read ``one"), the multiplicative truth, represents a set of observations whose size can be estimated probabilistically based on independence assumptions (e.g. the "optimal universe size" formula from PLN)
\item $ \bot $(read ``bottom"), the multiplicative falsity, represents the class of propositions grounded by no evidence
 \item $\mathbf{0} $(read ``zero"), the additive falsity, represents again the class of propositions grounded by no evidence
 \end{itemize}

 Intuitionistic and constructive logic began when people saw the possibility of reading $A \rightarrow B$ as ``if you give me an A, I will give you a B'', which is a significant departure from the classical reading ``B is true whenever A is.''   Specifically, the way to read $A \rightarrow B$ in the current interpretation/construction of linear logic is as: The evidential observations being used in favor of $A$, can be transferred to $B$ so they can be used in favor of $B$. That is: $A \rightarrow B$  means that, using the evidence supporting $A$ and the rules of logic (and accepting the heuristic assumptions underlying each rule application we do), we can conclude $B$.   

When we have a proof of $A \rightarrow B$ and a proof of $A$ in linear logic, by composing them we actually consume them to get a proof of $B$, so that $A \rightarrow B$ and $A$ are no longer available after the composition.    What this means is that: in this application of Modus Ponens, the pieces of evidence used to estimate the truth value of $A$ are used to help estimate the truth value of $B$.  Thus it is not OK to then use the the truth value of $A$ and the truth value of $B$ together in further inferences -- because this would entail double-counting of evidence.   In linear logic one says "A was consumed"; in this evidential interpretation we would say that "the observations used as evidence for A have been consumed in this particular inference and thus shouldn't be used again."

\begin{figure}
  \includegraphics[width=14cm]{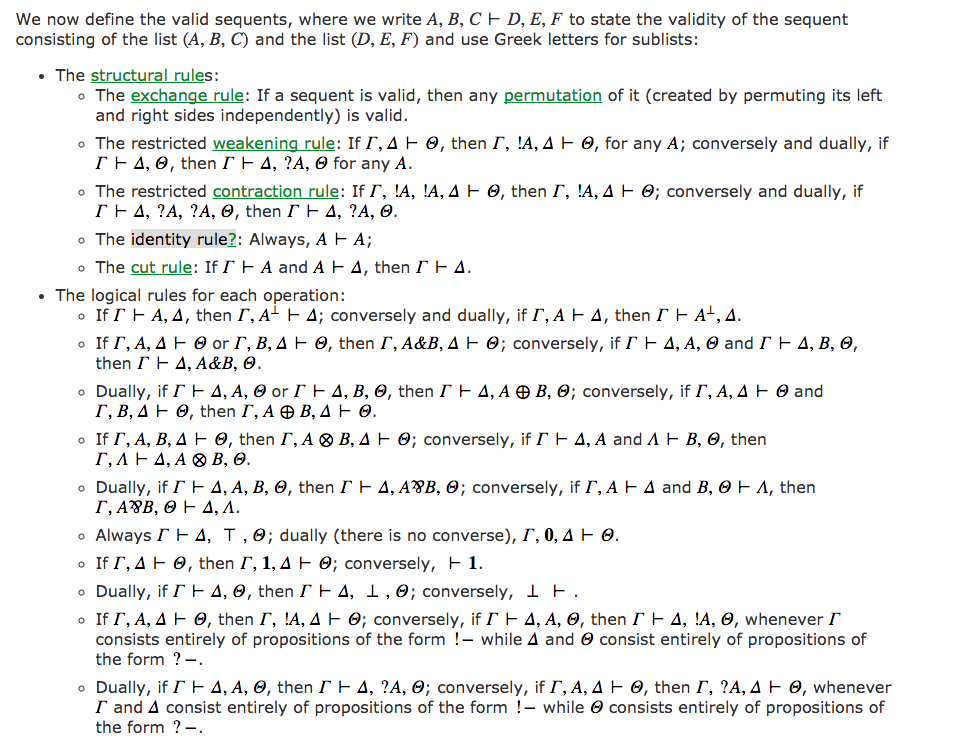}
  \caption{\label{fig:linear-logic-rules}Rules of linear logic}
\end{figure}

It is straightforward to verify that the structural and logical rules of linear logic (see \ref{fig:linear-logic-rules}) all hold in this interpretation, with appropriate uncertain truth value formulas associated to them.

However, an interesting complexity is that when we include the uncertain truth value formulas, we get a noncommutative variant of linear logic.   

The positive exponential ! has a fairly satisfying interpretation in terms of the standard resource interpretation of linear logic. Given a resource a, we know that $!a$ means an infinite supply of $a$. Or, stated more concretely in terms of the connectives of linear logic: $!a= !a \otimes a$.

In our evidential linear logic, $!a$ can be interpreted to mean an infinite amount of evidence about A.   So this means $n(A) = \infty$.  

Given that $?A=(!(A^\bot))^\bot$, we can also say that $?A$ means: The complement of the proposition stating there is an infinite amount of evidence in favor of the complement of A.   I.e., $?A$ means "A is still possible," which implies that either $n(A) < \infty$ or $p(A) > 0$.   

The standard rules for linear logic exponentials --

\begin{itemize}
\item If  $\Gamma ,A, \Delta \implies \Theta$ , then $ \Gamma ,!A, \Delta \implies \Theta$ ;  conversely, if  $\Gamma  \implies \Delta,A, \Theta$, then  $\Gamma  \implies \Delta,!A, \Theta$, whenever  $\Gamma$ consists entirely of propositions of the form !x  while  $\Delta$ and $\Theta$ consist entirely of propositions of the form ?x
\item Dually, if  $\Gamma  \implies \Delta,A, \Theta$ , then  $\Gamma  \implies \Delta,?A, \Theta$; conversely, if  $\Gamma ,A, \Delta \implies \Theta$ , then  $\Gamma ,?A, \Delta \implies \Theta$ , whenever  $\Gamma$ and  $\Delta$ consist entirely of propositions of the form !x  , while $\Theta$ consists entirely of propositions of the form ?x .
\end{itemize}

\noindent -- follow if one assigns $\Gamma  \implies \Delta,!A, \Theta$ and $\Gamma ,?A, \Delta \implies \Theta$ probabilistic truth values, although in this interpretation ! and ? turn out to be noncommutative as well (when one accounts for the associated quantitative truth value functions).

Exponential isomorphism, i.e.

$$
!(A \& B) \equiv !A \otimes !B 
$$

\noindent follows from the observation that if there is infinite evidence for the set of evidence in favor of both $A$ and $B$, under the assumption of maximal overlap of evidence,  then there must be infinite evidence obtainable by independently choosing from the infinite evidence for $A$ and the infinite evidence for $B$..

In linear logic, de Morgan's laws hold even for mixtures of additive and multiplicative operators.   Multiplication distributes over addition in this mixed sense, as follows:

\begin{enumerate}
\item $A \otimes (B \oplus C) \equiv (A \otimes B) \oplus (A \otimes C)$ (and on the other side);
\item $A \mid (B \& C) \equiv (A \mid B) \& (A \mid C)$ (and on the other side);
\end{enumerate}

\noindent This also works for the operations as defined here.    To validate the first equivalence, observe that

$$
a * max(b,c) = max(a*b, a*c)
$$

\noindent To validate the second equivalence, observe that

$$
a + min(b,c) - a * min(b,c) = min(a + b -ab, a+c -ac) 
$$

\noindent (since we assume $0 \leq a \leq 1$).

\section{Linear Logic via Fibring of Max Overlap and Probabilistic Uncertain Logics}

Putting these pieces together, we see that the rules of linear logic can be obtained via:

\begin{itemize}
\item (categorially) fibring together the and-or-not logic of uncertain propositions with $(s,n)$ truth values and a Max Overlap heuristic, with the and-or-not logic of uncertain propositions with $(s,n)$ truth values and a probabilistic-independence heuristic
\item Identifying the negation and the zero of the two logics being fibred together
\end{itemize}

Linear logic, as an abstract structure, can certainly be used to model many situations beside this one.  But this is one way of grounding the linear logic operations which is particularly intuitive, and is relevant to artificial intelligence applications and cognitive modeling.

\section{Linear Logic and Fuzziness}

Suppose that, following \cite{Goertzel2010e}, we model fuzzy characters like "tall" and "crazy", via considering there as being "tall detectors" and "crazy detectors" that we can apply to people. 

So if Ben is tall to degree .7 and crazy to degree .8, this means that when we hold the tall detector up next to Ben it rings 70\% of the percentage of time it can ring for anyone, and when we hold the crazy detector up next to Ben it rings 80\% of the percentage of time it can ring for anyone.

What, then is the degree to which Ben is "tall and crazy"?   There are two meanings here

\begin{enumerate}
\item One is holding a "tall and crazy" detector up to Ben, which rings only when Ben is judged both tall and crazy
\item One assumes that Ben is having a tall detector and a crazy detector held up simultaneously next to him, and wants to estimate the odds that, at any point in time during this experiment, both the tall detector and the crazy detector are ringing
\end{enumerate}

In the first case, if one assumes that the max possible amount of ringing is the same for both the tall and crazy detector, then one arrives at the conjunction formula

$$
B_{T \wedge C} = min(B_T, B_C)
$$

\noindent (where $B_T$ denotes the degree to which Ben is tall, etc.).

In the second case, if one assumes that the tall detector and the crazy detector are both configured to do all their ringing during the same interval of time after their initial placement next to Ben, and that the timing of each of their rings during that interval is random, and that the maximum activity would be ringing ceaselessly throughout the whole time interval -- then one arrives at the conjunction formula

$$
B_{T \wedge C} = B_T * B_C
$$

So here the two formulas for conjunctively combining fuzzy degrees are seen to represent different conceptions of the semantics of conjunction.

The count $n$ in this context may be interpreted as the number of times that the degree of output of the detector for a certain fuzzy character of a certain individual has been assessed.   

The formulas of linear logic would seem to apply to this case just as well as to the probabilistic logic case considered above.   The conclusion one comes to here is that fuzzy logic can be modeled as a special case of probabilistic logic, where one is looking at probabilities emanating from a "property detector" whose behavior one only observes in the aggregate (the amount of "ringing" it gives when exposed to a given stimulus) rather than at the micro level (one can't detect or remember the individual "rings").

(This also hints at some of the connections between fuzzy logic and quantum logic.   In the model of fuzzy logic suggested here, we are assuming that the individual rings are unobservable, and we are then making heuristic assumptions about their relationships.   But quantum logic connects in subtle ways with issues about what is and is not observable in principle.  But I'll leave it for later to unpack these connections...)





\bibliographystyle{alpha}
\bibliography{bbm}

\end{document}